# AGI Maze Prediction Datasets: A Compact Benchmark for Learning World Dynamics with Transformers

Alexey Potapov

SingularityNET Foundation
alexey@singularitynet.io

## Abstract

World modeling requires a predictive model to maintain and update an internal state adequate for reasoning about the consequences of actions. We introduce the **AGI Maze Prediction Datasets** and Benchmark, a lightweight controlled testbed for studying this capability in Transformers and other predictive models. Derived from procedurally generated, stateful grid worlds, the benchmark comprises per-step transition prediction, fixed-horizon state prediction, and sequential textual-observation prediction. Source-maze-disjoint training and validation splits, together with greedy exact-match evaluation, distinguish learning transferable action-conditioned dynamics from memorizing transitions in familiar layouts.

We establish from-scratch byte-level Transformer baselines and compare them with two working-memory-augmented architectures. A generic auxiliary latent-memory Transformer can fit some training sets perfectly but does not consistently improve held-out performance. In contrast, a pseudo-video spatial-memory Transformer initializes a two-dimensional latent workspace from the input map and updates it from action history without receiving intermediate maps, positions, or state labels. Under the same data, objectives, and evaluation protocol, this model reaches perfect validation accuracy on selected fixed-horizon tasks where the byte and unstructured-memory baselines do not, and substantially improves sequential text-trace prediction. These results suggest that structured, task-aligned working memory can be more useful than additional latent capacity alone. More broadly, we argue that language grounding is mediated by persistent data structures and computations over them; the benchmark offers a compact setting for testing architectures that couple textual interfaces to learned structured state.

## 1. Introduction

A central open problem on the path to general-purpose artificial intelligence is **world modeling**: learning internal representations of an environment that support prediction, state estimation, and action-conditioned reasoning. A useful world model must do more than associate a visible input with an immediate output. It must preserve information about the current state, update that information after an action, and make it available for predicting what will happen next. Such representations are central to model-based reinforcement learning and have enabled agents to learn compressed spatial and temporal models of environments, imagine possible futures, and improve behavior through those imagined trajectories [1, 2]. Learning structured state representations is particularly important when environments contain objects, relations, spatial topology, or other compositional regularities [3].

For large language models (LLMs), the problem takes a distinctive form. The standard LLM is trained and used primarily as a next-token predictor over a finite textual context. In an interactive task, its persistent state is therefore usually external to the network: previous observations are appended to the prompt; retrieved documents are inserted into it; and, when an agent is allowed tools, maps, notes, programs, or databases are maintained outside the model. This externalization is often useful, and executable world models maintained by coding agents are a promising practical approach [4]. Yet it does not establish that the predictive model itself has learned a stable, manipulable representation of the environment. In particular, a program placed in an agent workspace is an external artifact: it is constructed, inspected, and executed outside the forward pass, effectively as a black box from the perspective of the frozen language model.

This distinction matters because a textual action-observation history is not generally a natural representation of a structured state. Even a small grid world has two-dimensional adjacency, obstacles, object locations, inventory variables, and possibly non-local transition mechanisms. An LLM agent can describe these facts in prose, but the description is linear, costly to update, and prone to omission or inconsistency. A typical external note produced by SOTA LLMs with a world-modeling agentic harness while solving a maze may look as follows:

```
### World Model
* **Grid Size:** 4x4 (Rows 0-3, Cols 0-3)
* **Current Position:** (0, 1) - *Derived from: Start (3,3) -> Up (2,3)
  -> Left (2,2) -> Down (Blocked, stayed at 2,2) -> Up (River at 1,2,
  swept to 0,2) -> Left (0,1).*
* **Inventory:** {"key": false, "treasure": false, "exit_key": true}
* **Known Locations:**
  * (2, 1): Chest (Needs key)
  * (0, 1): Exit Key (Collected)
  * (1, 2): River (Sweeps to 0, 2)
  * (3, 2): Wall (Blocked movement)
```

The note can be helpful, but it is neither an intrinsic two-dimensional data structure nor a guarantee that the underlying model has a corresponding internal state. It is a serialized, externally maintained reconstruction of one. More generally, we hypothesize that robust world models should be able to maintain latent state in task-appropriate representational forms – potentially arbitrary data structures – and carry out computations directly over those structures. Rather than replacing natural-language descriptions of state, such representations provide their semantic grounding: they supply a stable internal basis for generating, interpreting, and validating textual descriptions of state. The two-dimensional spatial maps used in the present study are only one illustrative example of this broader principle.

In this respect, many current vision-language-action (VLA) models [5], which encode visual input as a sequence of projected visual embeddings appended to a language-model context, remain limited. Although these embeddings can retain information about the scene, the architecture does not in itself preserve the scene as an explicit, structured, and dynamically updateable spatial state on which task-relevant computations can be performed. We therefore view language-conditioned visual embeddings as a useful interface to perception and action, but not by themselves as a sufficient computational substrate for grounded spatial reasoning.

In our previous work, we introduced the **AGI Maze Benchmark Framework** [6], a lightweight family of interactive grid-world environments designed to isolate this issue without the confounds of pixel-level perception or continuous control. AGI Maze tasks are stateful and partially observable. An agent must infer position and map topology from action-observation pairs, remember discovered objects and inventory, and plan under a limited step budget. The basic environment includes walls, a border ("monolith") with a hidden exit, keys, and treasure; extensions add dynamics such as rivers that force movement and pits that teleport the agent through a cycle. These mechanics make the state-transition function depend on spatial configuration and cause simple local rules or unstructured action histories to become insufficient.

The earlier study found that contemporary LLMs and simple agents built on top of them do not solve even small instances reliably at a human level. Prompt-based working memory can improve results: agents often write coordinates, inferred objects, and future plans into their message history. However, the resulting representations remain external, textual, and brittle. These findings motivate a complementary question that cannot be answered by evaluating pretrained LLM agents alone: **can a predictive neural model trained from scratch learn an internal representation that supports the dynamics of a structured environment?**

We introduce the **AGI Maze Prediction Datasets** and the associated **AGI Maze Prediction Benchmark**[1] as a compact testbed for this question. The benchmark repurposes the AGI Maze framework for supervised predictive learning rather than interactive solving. It is intended for Transformers and other sequence or state-space models trained from random initialization – not for prompting pretrained LLMs. In this respect, its role is analogous to that of MNIST [7]: it is deliberately small, reproducible, and simple enough to support rapid controlled experiments, while retaining a sharply defined capability of interest. Here that capability is not visual classification, but learning action-conditioned state transitions, multi-step composition, and the transfer of transition rules across related two-dimensional worlds.

The datasets define three progressively more demanding prediction regimes. In the **per-step** task, a model receives a rendered maze state and one action, and predicts the next agent position. In the fixed-horizon **sequence** task, it receives the initial maze and a sequence of actions but no intermediate states; it must predict the final position after composing all transitions. Finally, in the **text-trace** task, it predicts the textual observations generated along an action trajectory. The latter formulation directly connects the benchmark to next-token prediction: the model must predict the consequences of actions in a compact language, not merely continue unrestricted natural text. Training and validation mazes are separated by source maze, preventing evaluation from being reduced to memorizing correlated transitions from the same layouts.

Our initial baselines demonstrate both the utility of the benchmark and the representational question it exposes. A small decoder-only **byte-Transformer**, trained from scratch on raw UTF-8 serializations of ASCII maps and action tags, can learn elementary transitions. However, its sample efficiency and exact predictive accuracy degrade as maze size, horizon, and transition complexity increase. In contrast, our **visual-memory Transformer** maintains a dedicated spatial working memory that is better suited to representing the state of the environment and the operations performed on it. It reaches 100% exact-target accuracy on several tests where a conventional byte-Transformer remains noticeably below 100%, which is diagnostically important: failure to achieve perfect accuracy indicates that the model has not fully learned the general rules governing the environment, and may instead be relying on memorized local transitions from the training data. The gap is still preliminary evidence rather than a proof of any particular mechanism, but it motivates controlled comparisons between sequential and explicitly spatial latent architectures.

[1] https://github.com/Necr0x0Der/agimaze-predict

The benchmark is designed to make such comparisons systematic. It supports ablations of representation format, recurrence or state slots, supervision targets, training-set composition, horizon, maze size, and mechanics. It also makes transfer questions concrete: whether examples from smaller mazes, different horizons, or simpler mechanics improve prediction in a target environment, and whether a model learns reusable transition rules rather than a collection of local correlations. Because the worlds are small, generated procedurally, and specified exactly, incorrect predictions can be inspected at the level of an individual state, action, and transition.

The contribution of this paper is therefore threefold:

1. We introduce AGI Maze Prediction Datasets, a family of controlled supervised datasets derived from AGI Maze for per-step transition prediction, multi-step state prediction, and sequential textual-observation prediction.
2. We formulate AGI Maze Prediction as a compact benchmark for probing whether predictive architectures learn representations that support structured, action-conditioned world dynamics and transfer across related environments.
3. We provide reproducible from-scratch baselines, including a byte-level decoder-only Transformer and a spatial visual-memory Transformer, establishing initial accuracy, sample-efficiency, and representation-format comparisons.

The goal is not to claim that success on small mazes constitutes general world modeling. Rather, AGI Maze Prediction provides a low-cost, diagnostically transparent setting in which a necessary component of world modeling can be measured: the construction and use of internal state sufficient to predict the consequences of actions in structured environments. It can serve as an early benchmark – or a "MNIST for world-dynamics prediction" – for testing ideas before they are scaled to visually rich, physically complex, or open-ended domains.

More broadly, AGI Maze Prediction can serve as a bridge between the world-modeling problem for language-based agents and latent predictive world models in visual control [8]. Its environments are presented in a simple symbolic and textual form, which makes the relation between state descriptions and internal representations transparent. Yet the central challenge – learning an action-conditioned, spatially organized latent state that supports prediction – is structurally the same as in visual world-modeling approaches such as LeWorldModel. This intermediate setting therefore permits controlled tests of representational hypotheses before the additional challenges of high-dimensional perception, continuous dynamics, and real-world control are introduced.

# 2. Datasets and Task Formulation

## 2.1 Lightweight Diagnostic Datasets for Predictive Models

Small, controlled datasets have long played an important role in the study of predictive models. Character-level corpora such as Tiny Shakespeare [9] are widely used as inexpensive sanity checks for autoregressive and next-token-prediction training. Synthetic tasks involving arithmetic, formal languages, copying, sorting, and modular operations provide more targeted tests of generalization and algorithmic behavior [10–12]. Such datasets are useful not because they approximate the full complexity of natural language, perception, or embodied interaction, but because they isolate a capability of interest under conditions in which training dynamics, failure modes, and out-of-distribution generalization can be measured precisely.

These diagnostic settings are not tied to a particular architecture: they can be used to compare Transformers with recurrent, state-space, memory-augmented, or other predictive models. They make it possible to ask whether a model has learned a rule or procedure rather than merely fit correlations in its training distribution, often using exact-match evaluation and deliberately constructed generalization splits [10–12]. Their small scale also permits rapid ablations of model capacity, representation format, supervision, data composition, and optimization choices.

AGI Maze Prediction belongs to this diagnostic tradition, but focuses on a different capability: learning an action-conditioned model of a persistent, structured environment. Unlike a task whose target follows from a static input or a purely one-dimensional symbolic rule, each example requires the model to represent a world state, update that state under one or more actions, and predict the resulting state or observation. The worlds remain deliberately small and procedurally generated, while source-maze-disjoint splits make it possible to distinguish learning reusable transition rules from memorization of local patterns in familiar layouts.

## 2.2 AGI Maze Prediction Datasets

All datasets are derived from procedurally generated AGI Maze environments [6] and are stored as UTF-8 JSON Lines files. A maze is rendered as an ASCII map whose walls, cells, objects, and current agent position are explicit. The current release contains maze families with keys on 3x3, 4x4, and 4x5 boards, and with rivers on 3x3 and 3x4 boards. These families vary both spatial scale and transition mechanics: keys introduce state-dependent object interactions, whereas rivers introduce forced, multicellular movement. Examples are collected from random action traces, with actions drawn from the four cardinal directions. Training and validation data are constructed from disjoint source mazes; thus, a validation example may follow a familiar local rule, but it is never drawn from exactly the same rendered layout as a training example. The validation splits support model selection and controlled comparison rather than serving as a final hidden test set.

The release comprises three complementary dataset formats. The first two share a compact map-and-action contract and use an exact position target. The third uses complete trajectories and requires autoregressive prediction of textual environmental observations. Together, they separate local transition learning, multi-step state composition, and language-level prediction grounded in an evolving world state.

### 2.2.1 Per-Step Transition Prediction

The **per-step** datasets isolate a single state transition. Each record contains a rendered current map and one action, expressed as <MAP>...</MAP> followed by <ACT>action</ACT>, and the target is the resulting zero-based agent coordinate, <POS>(row, column)</POS>. The map includes the agent's current position and all observable state needed for the requested transition. The task can therefore be written as

$$MAP_t, ACT_t \text{ -> } POS_{t+1},$$

for example,

```
{"input":"<MAP>\n+---+---+---+---+\n| K |\n+ + +---+ +\n| T | | \n+---
-+  +---+  +\n|  |  |  |\n+  +  +  +  +\n|  S  |\n+---+---+---+---
+\n</MAP>\n<ACT>right</ACT>","target":"<POS>(3, 2)</POS>"}
```

that corresponds to the map (Fig. 1).

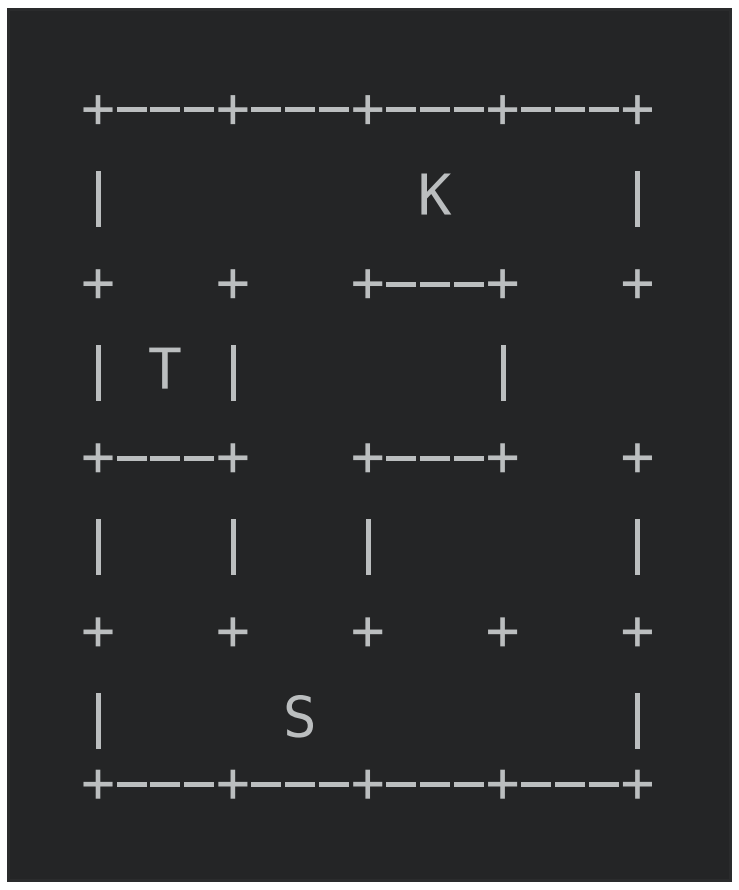

Figure 1. 4x4 maze example: S – start (player), K – key, T – treasure (see [6] for more details)

In contrast to the original AGI Maze Benchmark, it is not an agentic navigation or planning task: its purpose is to test whether a model can infer the local geometry and mechanics of a state transition from the representation supplied at that step. Random traces yield multiple examples per source maze. The current per-step collection contains 19,993 examples across five maze families, with training sets ranging from 1,875 to 6,758 examples and held-out validation sets ranging from 480 to 617 examples. Multiple training shards are provided for selected larger families, allowing data-scale and mixed-family experiments without changing the task contract.

### 2.2.2 Fixed-Horizon Sequence State Prediction

The **sequence** datasets test composition of transitions over a fixed action horizon. Each record again begins with the initial rendered map, but is followed by a sequence of action blocks; the target is the position after the last action: $MAP_1, ACT_1 \ldots ACT_h \text{-> } POS_{h+1}$.

No intermediate map, position, or observation is supplied after the initial state. A successful predictor must therefore preserve or reconstruct the relevant state across the complete action sequence, including the effects of walls, object interactions, and non-local mechanics. The released sequence datasets cover horizons of 1, 2, 4, and 8 actions for the 3x3-keys, 3x4-rivers, and 4x4-keys families. For each family-horizon combination, the release provides 4,000 training examples in two shards and 1,000 source-maze-disjoint validation examples. Trajectories are intentionally not deduplicated, so the data reflect the distribution of states and action subsequences encountered under the collection policy.

The per-step datasets can be viewed as the special case of the seq formulation with `h=1`. We nevertheless introduce them separately in order to isolate the learning of direct local transitions: given the current map and one action, the model only has to predict the next agent position, without the additional requirement to compose state updates across multiple steps.

The **sequence** formulation poses a stronger challenge. Given only the initial map and a sequence of actions, with no intermediate maps or positions, a model must predict the final state after the complete trajectory. To generalize systematically, it must implicitly learn and compose the relevant transition rules across the action sequence. An alternative is to fit direct associations between complete action sequences, maps, and endpoint positions, but this strategy scales poorly: the number of possible action sequences grows rapidly with the horizon, and their outcomes depend on the particular maze topology and mechanics.

### 2.2.3 Sequential Text-Observation Prediction

The **text-trace** datasets connect the benchmark directly to textual next-token prediction. Each JSONL record contains one complete trajectory in the form

```
<MAP>...</MAP>
<START>...</START>
<ACT>action_1</ACT>
<TXT>observation_1</TXT>
...
<ACT>action_n</ACT>
<TXT>observation_n</TXT>
```

The `<START>` block states the initial board size and agent coordinate; each `<TXT>` block is the environment's textual observation after the preceding action, for example a successful move, a blocked move, or an interaction with an object, e.g.

```
You went up and stepped on an empty tile.
You tried to go up, but a monolith blocks the way.
You tried to go right, but a wall blocks the way.
You went left and stepped on a tile with a chest. But you don't have a key.
You went up and stepped on a tile with a key. You pick up the key.
```

Models are trained to predict the bytes of the `<TXT>` blocks while the map or start description, action blocks, and earlier observations serve as context. For a later step in a depth-limited rollout, the preceding ground-truth observation is retained in the context. This teacher-forced evaluation prevents an early decoding error from being confounded with the model's prediction of subsequent transitions, while still requiring each target observation to be conditioned on the evolving trace.

This setup differs fundamentally from the **sequence** formulation. In the sequence task, intermediate positions are deliberately withheld: providing the ground-truth position after each action would decompose the problem into a series of independent one-step predictions. In the **text-trace** task, by contrast, the intermediate <TXT> blocks are observations available to an agent, not ground-truth descriptions of the full environment state. They may report the consequence of an action, but do not explicitly provide the agent's updated position, the complete map topology, or the latent variables required to determine subsequent transitions. The model must therefore learn to infer and maintain those intermediate states implicitly from the action-observation history. Teacher forcing of prior observations evaluates conditional next-observation prediction without compounding decoding errors; it does not supply the hidden world state that the model must learn to represent.

The current text-trace release contains 4,000 training trajectories and 1,000 validation trajectories from the 3x3-keys family. Complete traces contain 10 to 50 action-observation pairs; experiments may select either the map or the start description as the initial context and a maximum rollout depth. This format tests whether a model can use textual interaction history to predict the linguistic consequences of actions, while retaining the underlying requirement to track a structured environment state.

# 3. Baseline Models and Initial Experiments

## 3.1 Experimental Protocol and Evaluation

All baseline models are trained from random initialization on the explicit training shards described in Section 2. The corresponding validation shards are constructed from disjoint source mazes and are used for model selection and comparison. We report diagnostic experiments rather than an exhaustive, parameter-matched benchmark sweep: model width, depth, regularization, and training duration are kept small and are adjusted only as needed for the relevant dataset family. The purpose is to establish which task regimes can be learned reliably by simple predictive architectures, and where failures remain as board size, mechanics, or action horizon increase.

For the **per-step** and **sequence** tasks, the principal metric is greedy exact-target accuracy. At validation time, the model autoregressively generates the complete <POS>...</POS> target after the map-and-action prefix; a prediction is correct only if the entire target span matches the ground truth byte-for-byte. This exact metric is deliberately stricter and more interpretable than per-byte accuracy: an incorrect coordinate or malformed output counts as a failed state prediction. Target-byte negative log-likelihood is the training loss, computed only over unmasked target bytes; it is also retained as a diagnostic metric alongside greedy exact-target accuracy.

For **text-trace** tasks, loss is computed only on <TXT>...</TXT> spans. We report exact greedy accuracy both for individual observation spans and for complete depth-limited trajectories, where every evaluated observation must be generated exactly. As described in Section 2.2.3, earlier ground-truth observations are retained as context during rollout evaluation. Thus, these metrics measure conditional prediction of each next environmental observation without conflating it with accumulated errors from previous decoding steps.

## 3.2 Byte-Transformer Baseline

Our first baseline is a small decoder-only, GPT-style **byte-Transformer**. It uses literal UTF-8 bytes as its input vocabulary, learned positional embeddings, causal self-attention, pre-normalized residual blocks with GELU feed-forward layers, and tied input/output byte embeddings. The model receives the serialized map and action blocks as a single causal sequence. During training, next-byte cross-entropy is masked everywhere except the target span: `<POS>...</POS>` for per-step and sequence tasks, and `<TXT>...</TXT>` blocks for text traces. This setup tests whether a conventional autoregressive model can learn the required dynamics directly from the same textual representation that would be available to a language model, without an explicitly structured state representation.

The baseline establishes that the datasets are learnable in simple regimes. Table 1 reports representative **per-step** results. The 3x3-keys task is solved perfectly, whereas comparable one-step datasets already reveal sensitivity to board size and transition mechanics. Additional data largely closes the gap for 4x5-keys, but the more complex 3x4-rivers family remains below perfect exact accuracy even when the training set is enlarged and supplemented with related maze families.

All byte-Transformer configurations use four attention heads and an MLP expansion factor of four. In Tables 1 and 2, `C` denotes context length, `d` model width, `L` the number of Transformer layers, and `p` dropout probability. Full training configurations are provided in the accompanying experiment files.

**Table 1.** Byte-Transformer results on **per-step** transition prediction. Accuracy is greedy exact-target accuracy on the held-out validation split.

| **Maze family** | **Training set** | **Training examples** | **Model configuration** (`C d L p`) | **Validation accuracy** |
|---|---|---|---|---|
| 3x3-keys | 1 shard | 2,211 | `512 128 4 0.00` | 1.0000 |
| 4x4-keys | 1 shard | 2,364 | `512 256 4 0.04` | 0.9887 |
| 4x5-keys | 1 shard | 2,215 | `512 512 4 0.04` | 0.8253 |
| 4x5-keys | 2 shards | 4,489 | `512 512 4 0.04` | 0.9981 |
| 4x5-keys | 3 shards | 6,758 | `512 512 4 0.04` | 1.0000 |
| 3x3-rivers | 1 shard | 1,875 | `512 256 4 0.00` | 0.9833 |
| 3x4-rivers | 1 shard | 2,051 | `512 256 4 0.00` | 0.8321 |
| 3x4-rivers | 2 shards | 4,058 | `512 256 4 0.00` | 0.9354 |
| 3x4-rivers | 5 shards (mixed) | 10,508 | `512 256 4 0.00` | 0.9557 |

The experiment in the last line in the Table 1 mixed two training shards of 3x4-rivers with 3 different shards – 3x3-keys, 4x4-keys, 3x3-rivers. Interestingly, it increased the score by 2% implying modest but non-zero transfer of information from different but related families of mazes.

Although the Transformer may generalize beyond direct memorization of individual local transitions to a limited extent, examples from mazes of other sizes provide only modest benefit. Increasing the maze size from 3x3 to 4x4 and 4x5 requires several times more training data, even though the underlying transition laws are identical. This suggests that the model does not learn these general rules in a robust, reusable form, possibly because its standard sequential representation does not provide a computationally suitable means of instantiating and applying them over the structured state of a particular maze.

Fixed-horizon **sequence** prediction amplifies this pattern. Table 2 shows that increasing the number and diversity of training examples improves the results, but does not uniformly recover perfect exact prediction. The 3x3-keys two-step task approaches perfect accuracy with 4,000 target-horizon examples, whereas longer horizons, larger mazes, and river mechanics remain substantially more demanding.

**Table 2.** Byte-Transformer results on fixed-horizon position prediction. Accuracy is greedy exact-target accuracy on the held-out validation split.

| Maze family | `h` | Training set | Training examples | Model configuration (`C d L p`) | Validation accuracy |
|---|---|---|---|---|---|
| 3x3-keys | 2 | 1 shard | 2,000 | `512 128 4 0.00` | 0.964 |
| 3x3-keys | 2 | 2 shards | 4,000 | `512 128 4 0.00` | 0.998 |
| 3x3-keys | 4 | 1 shard | 2,000 | `512 128 4 0.00` | 0.845 |
| 3x3-keys | 4 | 2 shards | 4,000 | `512 128 4 0.00` | 0.901 |
| 3x3-keys | 4 | 6 shards (mixed 2-, 4-, 8-step) | 12,000 | `512 128 4 0.00` | 0.979 |
| 4x4-keys | 2 | 1 shard | 2,000 | `512 128 5 0.05` | 0.726 |
| 4x4-keys | 2 | 2 shards | 4,000 | `512 128 5 0.05` | 0.894 |
| 4x4-keys | 2 | 10 shards (mixed horizons and related 3x3-keys shards) | 20,000 | `512 128 5 0.05` | 0.951 |
| 3x4-rivers | 2 | 10 shards (mixed horizons and 4x4-keys shards) | 20,000 | `512 128 5 0.05` | 0.881 |

The **text-trace** results likewise show that accurate byte-level prediction does not automatically yield perfect trajectories: each trace-level success requires every predicted observation to be exact. Both text-trace configurations use `C=2048`, `d=256`, `L=5`, `p=0.00`. For the 3x3-keys family:

- With 2,000 training trajectories and a rollout depth of four, the model reaches `txt_span_exact_accuracy=0.9105` and `trace_all_txt_exact_accuracy=0.701`.
- With 4,000 training trajectories and a rollout depth of eight, it reaches `txt_span_exact_accuracy=0.9809` and `trace_all_txt_exact_accuracy=0.869`.

Taken together, these results show that a conventional byte-level autoregressive model can fit simple transition regimes and improve substantially with additional or more diverse data. At the same time, the rapidly increasing data requirement and the persistence of exact-match errors on small, fully specified worlds suggest that the model often learns (or memorizes) an incomplete or ad hoc approximation to the transition function rather than a fully reusable representation of world state and mechanics. This interpretation is diagnostic rather than conclusive: the present experiments do not identify the internal mechanism of the byte-Transformer. They do, however, motivate the following comparison with architectures that provide an explicit latent state and, subsequently, a spatially organized one.

## 3.3 Auxiliary Latent-Memory Transformer

To separate the effect of an explicit spatial representation from the effect of simply adding latent capacity, we evaluated an **auxiliary latent-memory Transformer** on the **sequence** task. In addition to the byte-level autoregressive stream, the model maintains a set of learned working-memory cells processed by a second Transformer. The intended role of this auxiliary memory is to provide a persistent internal workspace in which the model can encode and update the environment state, rather than relying only on token embeddings optimized for next-byte prediction. Unlike the visual-memory architecture described in the next section, these cells have no predefined two-dimensional topology and receive no direct supervision over their contents.

Table 3 shows that the modified model systematically underperforms the byte-Transformer baseline; we therefore did not consider a more extensive experimental sweep necessary.

**Table 3.** Auxiliary latent-memory Transformer results on fixed-horizon position prediction. Accuracy is greedy exact-target accuracy on the held-out validation split.

| Maze family | h | Training set | Validation accuracy | byte-Transformer baseline |
|---|---|---|---|---|
| 3x3-keys | 4 | 6 shards (mixed) | 0.975 | **0.979** |
| 4x4-keys | 2 | 10 shards (mixed) | 0.886 | **0.951** |
| 3x4-rivers | 2 | 10 shards (mixed) | 0.868 | **0.881** |

It should be noted that the auxiliary model can completely fit at least some training regimes: for the 3x3-keys experiment it reaches target negative log-likelihood of zero and greedy exact-target accuracy of 1.000 on the training set. Its held-out accuracy nevertheless remains 0.975, slightly below the corresponding byte-Transformer result (0.979). The same pattern is clearer on 4x4-keys, where the auxiliary model reaches 0.886 compared with 0.951 for the byte-Transformer; on 3x4-rivers the difference is smaller (0.868 versus 0.881).

Thus, the additional Transformer and unconstrained memory cells are sufficient to increase fitting capacity, but do not by themselves yield a more useful or generalizable state representation. The training-validation gap is consistent with the auxiliary path adding parameters and computational routes that make optimization and overfitting easier, without supplying an inductive bias for storing the relevant environment state. This is not evidence that latent working memory is intrinsically unhelpful: with larger datasets or explicit learning signals directed at memory contents, it may become useful. In the present formulation, however, the model does not reliably learn to use the added cells as a state representation. This control is important for interpreting the visual-memory results below: their advantage cannot be attributed merely to extra latent memory, but depends on the task-aligned spatial organization and update mechanism.

## 3.4 Pseudo-Video Spatial-Memory Transformer

We next introduce a **pseudo-video spatial-memory Transformer**, which replaces the unstructured auxiliary cells with a persistent two-dimensional latent frame. The name emphasizes that the model is not trained on videos and is not given a video of the trajectory. The initial rendered ASCII map is placed, with its row-column geometry preserved, into the first frame of a fixed visual canvas. Character, row, and column embeddings are then transformed by spatial-attention blocks. Each completed action block produces a text-derived write context that updates the current latent frame; spatial blocks and causal per-cell temporal-attention blocks generate the subsequent frame. The resulting sequence of continuous frames can be viewed as an internally constructed "pseudo-video" or an evolving visual imagination of the maze.

The geometry of the *initial* frame is specified by the input representation, but the architecture is not told how to transform that frame after an action, what any latent cell should encode, or whether the generated frames should resemble rendered maps. All post-initial frames are continuous, learned states and receive no map reconstruction loss, coordinate labels, or intermediate-state supervision. In particular, no intermediate visual maps, positions, or ground-truth trajectory frames are supplied in the **sequence** task. The model may therefore learn to preserve and update a useful spatial state, or may learn to ignore this workspace; only the final position-prediction objective distinguishes these possibilities.

This comparison uses the same source examples, action sequences, train-validation splits, target-only next-byte NLL, and greedy exact-target metric as the byte-Transformer and auxiliary-memory baselines. It has no additional observations or labels. Its only relevant advantage over the auxiliary model is an inductive bias: the working memory has a native two-dimensional arrangement matching the spatial organization of the input maze.

We tested two readout variants, which differ in how the textual prediction pathway accesses the spatial memory. In the **full-text** variant, the source text is processed by a causal text Transformer; at each layer, each text-token representation first undergoes causal self-attention and then reads the visual frame current at that token's action-history position through cross-attention. The visual-memory stream is updated separately: after every completed block, a causal action encoder produces an action-context vector, which is written into the current two-dimensional latent frame and transformed by spatial and causal temporal self-attention. Thus, information flows from the action-text representation into the visual memory at discrete action events, and from the current visual memory back into the text stream through cross-attention.

In the stricter **visual-only** variant, position generation is performed by a separate causal target decoder. This decoder receives only the known query prefix and the previously generated target bytes through self-attention, while it reads the final visual frame through cross-attention. Here, the final visual frame denotes the two-dimensional latent state obtained by initializing the memory from the input map and then applying one learned update after each action in the requested sequence. The target decoder has no direct access to the serialized map or action sequence. Correct prediction therefore requires the relevant information from both the initial map and the action history to have been stored and transformed within the visual workspace. The results for the **sequence** experiment are summarized in Table 4.

**Table 4.** Fixed-horizon **sequence** results for the byte-Transformer baseline and the pseudo-video spatial-memory Transformer. Accuracy is greedy exact-target accuracy on the held-out validation split.

| Maze family | h | Training set | Byte-Transformer baseline | Full-text readout | Visual-only readout |
|---|---|---|---|---|---|
| 3x3-keys | 4 | 6 shards (mixed) | 0.979 | 0.978 | **1.000** |
| 4x4-keys | 2 | 10 shards (mixed) | 0.951 | 0.976 | **1.000** |
| 3x4-rivers | 2 | 10 shards (mixed) | 0.881 | 0.905 | **0.973** |

The visual-only model reaches perfect validation accuracy on both the 3x3-keys four-step and 4x4-keys two-step tasks, where neither the byte-Transformer nor the unstructured auxiliary-memory model is close to 1.000. It also greatly improves the most difficult 3x4-rivers result to 0.973, compared with 0.881 for the byte-Transformer and 0.868 for the auxiliary model. Because the visual-only decoder cannot directly access the serialized map or action sequence, these results show that the model has learned to use the auxiliary spatial stream in a way that supports held-out prediction; merely adding a second Transformer, as in Section 3.3, is not sufficient to obtain this behavior.

The same architecture also improves sequential textual-observation prediction. On the 3x3-keys **text-trace** task with an eight-step rollout, trained on 4,000 trajectories, the visual-only model reaches `txt_span_exact_accuracy=0.9996` and `trace_all_txt_exact_accuracy=0.997`, compared with 0.9809 and 0.869, respectively, for the byte-Transformer. The remaining errors involve predicting a chest interaction as occurring with or without a key. Such interactions are relatively rare within the first eight steps, so this error is consistent with limited coverage of that state-dependent event rather than a general failure to maintain the trajectory state. This improvement is also very prominent.

The result should nevertheless be interpreted carefully. Perfect exact-target accuracy on source-maze-disjoint validation sets is strong evidence that the architecture can learn transferable transition information under these conditions, but it does not prove that individual latent cells have a one-to-one, interpretable correspondence with maze cells or objects. Nor are the intermediate frames trained to reproduce the original rendered map: after initialization they are free to encode any continuous state useful for prediction. They may nonetheless contain interpretable spatial structure, which can be tested with probes of memory contents, ablations of the spatial topology and temporal update mechanism, or causal interventions on selected latent cells. The current experiments support the narrower conclusion that a two-dimensional, dynamically updated working-memory structure provides a useful inductive bias for this family of world-dynamics prediction tasks.

More broadly, the visual memory can serve as a grounding substrate for the text channel: it connects linguistic action and position/observation tokens to a persistent spatial state on which the model can compute. A natural next experiment is to train on mixed datasets containing both trajectories with intermediate maps and trajectories that provide only text descriptions, then test whether the model can construct useful two-dimensional states from text alone and, conversely, generate or validate text from those states. The objective is not to build a specialized solver for AGI Maze mechanics, but to test the more general architectural hypothesis that latent two-dimensional working memory can substantially extend Transformer capabilities for spatial reasoning. If this principle continues to hold under broader conditions, the same division between a textual interface and grounded spatial workspace could be incorporated into LLM architectures.

# Conclusion

We introduced the AGI Maze Prediction Datasets and Benchmark as a lightweight, controlled setting for studying world-modeling capabilities in Transformers and other predictive models. The benchmark turns stateful grid-world interaction into supervised per-step transition prediction, fixed-horizon state prediction, and sequential text-observation prediction. Its procedurally generated, source-maze-disjoint splits and exact-match metrics make it possible to evaluate whether a model transfers action-conditioned transition rules to previously unseen layouts, while retaining the low cost and diagnostic transparency needed for rapid architectural experiments.

The initial baselines demonstrate why this setting is useful. A conventional byte-level Transformer can learn simple transition regimes, but its accuracy declines as maze size, horizon, and mechanics become more demanding, and it may require substantial additional data to approach perfect exact prediction, which is expected from a capable model. Adding an unstructured auxiliary latent stream does not resolve this limitation: it can fit the training data perfectly while generalizing less well than the byte baseline. By contrast, the pseudo-video spatial-memory Transformer, trained and evaluated on the same examples and targets and given no intermediate maps or state labels, reaches perfect validation accuracy on selected **sequence** tasks and substantially improves **text-trace** prediction. These results do not establish a complete theory of its internal representation, but they show that the model can learn to use a two-dimensional, dynamically updated latent workspace in a way that improves held-out prediction.

The broader claim is not that a Transformer must directly ground language in raw external reality, nor that AGI Maze should produce a specialized solver for one set of game mechanics. We hypothesize that present LLM architectures do not reliably construct internal representations for arbitrary structured data, such as spatial maps, graphs, or other task-specific state structures. Instead, they commonly mediate such structures and the operations over them through textual descriptions that the model can process directly, or through external artefacts such as program code. This may allow an LLM to state a general rule – for example, that an agent remains in place when it attempts to cross a wall – but without internally maintaining the structured state needed to apply that rule reliably in a particular environment.

Grounding is therefore mediated by representations: language refers to a world through persistent data structures that encode relevant aspects of state and through computations that update, query, and relate those structures to text. In the present experiments, a latent two-dimensional visual memory supplies such a substrate for spatial state, while the text channel supplies actions, observations, and predictions. The improvement of the spatial-memory model suggests that this arrangement can be beneficial for predictive models. More generally, it motivates an architectural direction for language-based models in which a textual interface is complemented by learned, task-appropriate working memories on which structured state can be maintained and transformed.

AGI Maze Prediction is intended as a compact testbed for making this hypothesis falsifiable. Natural extensions include parameter-matched architecture comparisons, longer horizons, broader maze families and mechanics, partial observability, probes and causal interventions on latent memory, and mixed map-text training in which spatial states must be inferred from language or rendered back into it. Success on these small environments is not sufficient for general world modeling; nevertheless, the ability to construct and use an internal state adequate for exact action-conditioned prediction is a measurable prerequisite. The benchmark provides a practical way to study that prerequisite before scaling the same representational ideas to richer visual environments and to LLM architectures.

# References

1. Ha D., Schmidhuber J. *World Models*. arXiv:1803.10122, 2018. https://arxiv.org/abs/1803.10122
2. Hafner D., Pasukonis J., Ba J., Lillicrap T. *Mastering Diverse Domains through World Models*. arXiv:2301.04104, 2023. https://arxiv.org/abs/2301.04104
3. Kipf T., E. van der Pol, Welling M. *Contrastive Learning of Structured World Models*. ICLR, 2020. arXiv:1911.12247. https://arxiv.org/abs/1911.12247
4. Rodionov S. *Executable World Models for ARC-AGI-3 in the Era of Coding Agents*. arXiv:2605.05138, 2026. https://arxiv.org/abs/2605.05138
5. Brohan A. et al. *RT-2: Vision-Language-Action Models Transfer Web Knowledge to Robotic Control*. arXiv:2307.15818, 2023. https://arxiv.org/abs/2307.15818
6. Potapov A. *AGI Maze as a Benchmark Framework for World-Modeling Agents*. arXiv:2607.00627, 2026. https://arxiv.org/abs/2607.00627
7. LeCun Y., Cortes C., Burges C.J.C. *The MNIST Database of Handwritten Digits*. http://yann.lecun.com/exdb/mnist/
8. Maes L., Le Lidec Q., Scieur D., LeCun Y., Balestriero R. *LeWorldModel: Stable End-to-End Joint-Embedding Predictive Architecture from Pixels*. arXiv:2603.19312, 2026. https://arxiv.org/abs/2603.19312
9. Karpathy A. *nanoGPT: The Simplest, Fastest Repository for Training/Fine-Tuning Medium-Sized GPTs*. GitHub repository, 2022. https://github.com/karpathy/nanoGPT
10. Power A., Burda Y., Edwards H., Babuschkin I., Misra V. *Grokking: Generalization Beyond Overfitting on Small Algorithmic Datasets*. arXiv:2201.02177, 2022. https://arxiv.org/abs/2201.02177
11. Delétang G. et al. *Neural Networks and the Chomsky Hierarchy*. arXiv:2207.02098, 2022. https://arxiv.org/abs/2207.02098
12. Jelassi S., d'Ascoli S., Domingo-Enrich C., Wu Y., Li Y., Charton F. *Length Generalization in Arithmetic Transformers*. arXiv:2306.15400, 2023. https://arxiv.org/abs/2306.15400